\documentclass[twoside,11pt]{article}

\usepackage{jmlr2e}
\usepackage{mathptmx}
\usepackage{amsmath}
\usepackage{amsfonts}
\usepackage{soul}\setuldepth{article}
\usepackage[utf8]{inputenc}
\usepackage{graphicx}
\usepackage[noend]{algpseudocode}
\usepackage{algorithm}
\usepackage{enumitem}
\usepackage[table]{xcolor}
\usepackage{tabularx}

\ShortHeadings{DeepAL for Regression Using $\epsilon$-weighted Hybrid Query Strategy}{Vardhan and Sztipanovits}
\firstpageno{1}

\begin{document}
\title{Deep Active Learning for regression using $\epsilon$-weighted Hybrid Query strategy}
\author{\name Harsh Vardhan \email harsh.vardhan@vanderbilt.edu \\       \name Janos Sztipanovits \email janos.sztipanovits@vanderbilt.edu \\
       \addr Department of Computer Science\\
       Vanderbilt University\\
       Nashville, TN 98195-4322, USA}
       
\maketitle
\editor{}
\vspace*{-13mm}
\begin{abstract}
Designing an inexpensive approximate surrogate model that captures the salient features of an expensive high-fidelity behavior is a prevalent approach in design optimization. In recent times, Deep Learning (DL) models are being used as a promising surrogate computational model for engineering problems. 
However, the main challenge in creating a DL-based surrogate is to simulate/label a large number of design points, which is time-consuming for computationally costly and/or high dimensional engineering problems. In the present work, we propose a novel sampling technique by combining the active learning (AL) method with DL.
We call this method \textit{$\epsilon$-weighted hybrid query strategy ($\epsilon$-HQS)}, which focuses on the evaluation of the surrogate at each learning iteration and provides an estimate of the failure probability of the surrogate in the Design Space. By reusing
already collected training and test data, the learned failure probability guides the next iteration's sampling process to the region of the high probability of failure. 
 We empirically evaluated this method in two different engineering design domains, \textit{finite element based static stress analysis of submarine pressure vessel} (computationally costly process) and second \textit{submarine propeller design} ( high dimensional problem). Empirical evaluation shows better performance of our technique as compared to other methods and in one case was able to achieve same level of prediction accuracy as random sample selection process with 58\% less number of samples. 
\url{https://github.com/vardhah/epsilon_weighted_Hybrid_Query_Strategy}. 
\end{abstract}

\vspace*{-4.5mm}
\section{Problem Formulation}
\vspace*{-3mm}
\label{sec:problem}
In the context of this paper, Design Space Exploration (DSE) is defined as a process to characterize the response surface of a simulator's input--output behavior, where the input describes the relevant system parameters, and the output describes quantities of interest. These simulators can be parameterized and the values of a parameter can be changed in a specified bounded range. These all possible combinations of different parameter values construct the design space.   We will first introduce our notations for formalization. 
Let $DS$ be a bounded subspace of $\mathbb{R}^d$, i.e., $DS \subset \mathbb{R}^d $ and it is represented by an unlabeled set of $n$ samples. $d$ is the number of input parameters that describes the relevant system  properties. The training data is $D_{train}=(X,Y^*)$, where $X$ and  $Y^*$ are the possible collection of all $x_i $ (input samples to the simulator) and $y_i^*$ (corresponding simulation's output).  The role of a simulator is to map the input space to the solution space. i.e. $Sim:X \rightarrow Y^*$.  We view these simulators as a nonlinear function which can be modeled with a $k$--parameter configuration nonlinear regression model i.e. $Sim=Model(\theta_1,\theta_2,...,\theta_k)$. 
Accordingly, the surrogate modeling problem is the conditioning of a parametric model ($h$) on $D_{train}$ to find its parameters ($\theta$) which results into:  
$ h(x, \theta|D_{train}) =y, \; \forall \,x \in DS : y \approx y^* $ where $y^*$ represents the ground truth.
Since our evaluation is costly and/or has high dimensional design space, we only want to take strategic samples with the goal to obtain a certain level of prediction accuracy with the least amount of simulation queries. In this case, our problem can be formalized as:
\vspace*{-5mm}
\begin{gather*}
    min \; |{D_{train}}| \\ \textit{subject to} \;\;
    h(x,\theta|D)=y,\,  \forall \,x \in DS :  y\approx y^*  
\end{gather*}   
This is an optimization problem, whose direct solution is NP-hard \citep{cunningham1997combinatorial}. However, the heuristic-based iterative approach has been successful and this field of research is known as Active Learning \citep{settles2009active}.  The interpolant model $h$ is an instance of a computational architecture $A$, which defines the functional dependence of the interpolant on its parameter $\theta$.  We choose the architecture $A$ to be a Deep Neural Network(DNN), which has a proven track of powerful learning capability, cheap parallel prediction, and automatic feature extraction. Deployment of active learning strategies is tightly bounded with the class of chosen architecture. 
Active learning is a semi-supervised machine learning approach for sequential sampling with the goal of identifying strategic samples. The general flow of the AL procedures works as follows: 
\setlist{nolistsep}
    \begin{enumerate}[noitemsep]
    \item Select some initial randomised samples via methods like Uniform random,\/ Latin hypercube sampling and evaluate these to create $D^{train} \gets D_{train}^{init}$.
    \item Train a machine learning model ($h$) using $D^{train}$ for constraining the sampling procedure in next iteration.
    \item Design a \textit{query strategy ($Q$)} to compute a score ($s \in \mathbb{R^+}$) using the trained model ($h$) on entire design space ($DS$).
    \item Select new samples in next iteration based on score ($s$), and evaluate these samples to create new additional training data ($D_{train}^{new}$)
    \item Update $D^{train} : D^{train} \gets D_{train}^{new}$ 
    \item Iterate 2-5 until budget of training sample ($B=  |{D_{train}}| $)
\end{enumerate}

By formulating the AL framework in a pool-based sampling scheme, where initially the design space $DS$  has an unlabeled set of $n$ samples and the current labeled training set $D^{train}$ is the labeled set of $m$ samples, our goal is to design a query strategy $Q$ such that $Q: DS \mapsto D^{train} $. The combined goals in a supervised environment can be expressed as follows:
\vspace*{-1mm}
\begin{gather*}
    \underset{D^{train} \subseteq DS ,(x,y)\in D^{train}}{\mathrm{argmin}} \mathbb{E}_{(x,y)}[l(h(x,\theta|D^{train})] \\
   min \; |{D_{train}}| \\ \textit{subject to} \;\;
    h(x,\theta|D)=y,\,  \forall \,x \in DS :  y\approx y^* 
\end{gather*}
where $l(h(x,\theta|D^{train})$ is the loss function and $l(.) \in R^+$ and the first goal is to reduce loss on $D^{train}$ and the second goal is to reduce $m$ ($m=|{D_{train}}|$) as small as possible \textit{while ensuring a predetermined level of accuracy on $DS$}. Therefore, the query strategy $Q:DS \mapsto D^{train} $ is crucial. Also at each acquisition step, we want to select a batch of data points $\{x_1^{sel} ,x_2^{sel} ,...,x_b^{sel}\}$ rather than just one point $x^{sel}$ because retraining the learning model ($h$) on every single additive data-point would be computationally infeasible and even may not give any performance gain. So, based on our current trained model's state $h(w|D_{train})$, we need to design the query strategy ($Q$) which scores a candidate batch of unlabelled data points ($L_{candidate}$) from $DS$.
An Active learning framework has various query strategies like (uncertainty-based, variance-based, and disagreement-based) however directly applying these strategies to regression problems out of the Bayesian framework is less understood/studied. Accordingly, the standard AL approach is to estimate their uncertainty directly, using Bayesian methods \citep{rasmussen2003gaussian}, or estimate uncertainty
indirectly, through the variance of the outputs of an ensemble of models. Bayesian methods are computationally expensive when  the number of samples in DS is more than $10000$ and/ or input dimensions are more than $10$. Ensemble-like models require independent training of several models, which is also time-consuming due to training a family of models.
In this work, we tried to solve this problem out of the Bayesian framework and ensemble models to make it scalable and practically useful for a large class of problems. In the next section, we will explain our approach.   
 
\vspace*{-5mm}
\section{Approach}
\begin{figure}[h!]
        \centering
        \includegraphics[width=15.5 cm]{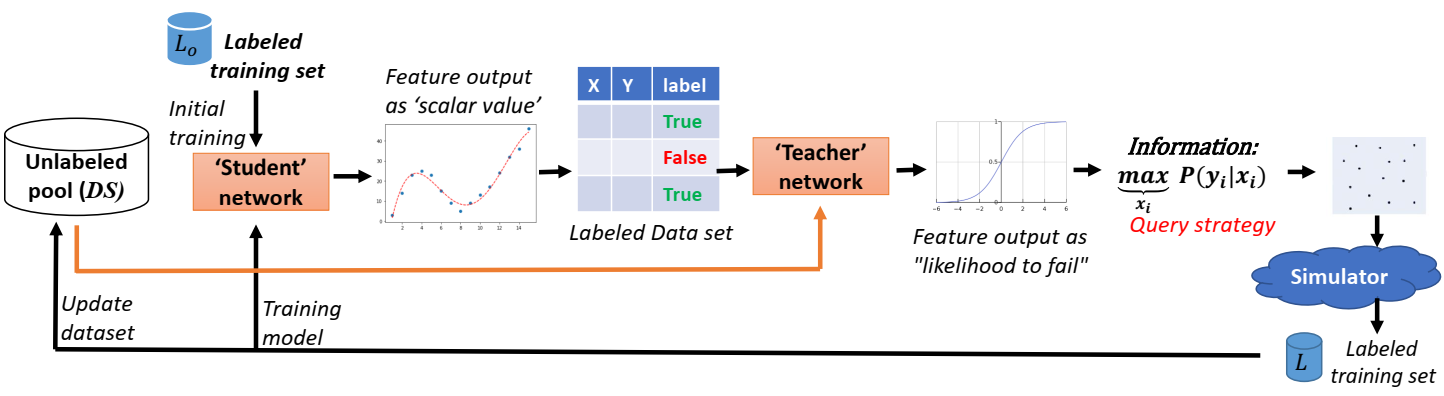}
        \vspace*{-9mm}
        \caption{Student teacher architecture: the process}
        \label{fig:st}
    \end{figure}

\vspace*{-2mm}
\label{sec:approach}
 To integrate the DL framework with the AL framework, we propose to use two neural networks in a \textit{student-teacher} fashion, first network is called the \textit{student} network (whose goal is to learn the input-output behavior) and the second network is called the \textit{teacher} network (whose goal is to guide next iteration's sampling process). For training the teacher network, we use already evaluated historical training and test data from prior iterations and label it with a binary indicator ($C$), where $C: X \mapsto \{0,1\}$, where $C = 1$, indicates student network is able to predict on the sample within an acceptable level of accuracy otherwise $C = 0$. The objective of training the teacher network is to make an inference on the failure probability of the student network on the entire design space. This failure probability can be used  to identify the regions/ samples in $DS$ on which student network has poor performance/ high likelihood to fail. We continue refining the student and teacher networks on further evaluated samples at each iteration  until we exhaust our budget for iterations. (refer figure \ref{fig:st}) 
\vspace*{-2mm}
Ideally, the true failure probability function $f^*$ is unknown but for sampling purposes, the approximation would work. The benefit of choosing a binary failure indicator is, that we can have a very small DNN because creating a coarse approximation of the decision boundary is a much simpler job than creating a nonlinear regression model. By leveraging the GPU implementation of DNN, we can make inferences on thousands/millions of samples at a cost negligible to running the costly simulation. With the knowledge of the failure probability of student network, it is possible to identify at each iteration, what are the samples $x \in DS$ on which student network has a high likelihood to fail.  
\vspace*{-4mm}
\subsection{Batching}
\vspace*{-2mm}
Using the learned probability function, the active learning framework ideally would select one sample per iteration which has \textbf{maximum probability to fail}. However, this one additive sample per iteration does not go well with the DL framework because of retraining time. So, at each iteration, a batch of data points ($L_{candidate}$) should be selected. For a given batch size ($B$) , the goal is to design an query strategy $Q$ such that $Q:L_{candidate} \rightarrow DS$. The most obvious approach for batching is to select top $B$ samples sorted in descending order by failure probability, we call this approach as \textbf{top-B}, where $B$ is the batch size at each iteration. However, this method is highly likely to select a set of information-rich but similar samples because the query strategy considers each sample independently and ignores the correlation between samples. This may result in a wastage of the evaluation budget. Our goal is not only to select information-rich samples but also to maintain diversity in the selected samples.  One way to solve it is by combining probability information with unsupervised clustering. \cite{zhdanov2019diverse} proposed the batched active learning using the clustering approach and it utilizes both the amount of information in the probability function as well as the diversity of information. We call this approach Diverse Batched Active Learning (\textbf{DBAL}).  
\vspace*{-4mm}
\subsection{Debiasing scheme }
\vspace*{-2mm}
Empirical evaluation of both \textbf{DBAL} and \textbf{top-B} sampling approaches performed poorer than randomized sampling. The underlying reason is, in the initial phase of training, the teacher network which learns the probability function ($f:x \rightarrow [0,1] \;\forall\; x \in DS$) has only limited information about the design space $DS$ (a maximum up to training and testing data). Accordingly, the \textit{``trained model does not know what it does not know"} and it creates a biased estimate of the probability function. If $f^*$ is the true failure probability function then $f$ can approximate it only up to $\beta-approximation$. The value of $\beta$ is an open research question, but we can increase the robustness of the teacher network and can reduce the sensitivity of mismatch between the distribution of $f$ and $f^*$. For this purpose, we introduce a hyper-parameter $\epsilon$ ($0 < \epsilon \leq 1$) and adopt filtering of evaluated samples using rejection threshold and then running $\epsilon$-weighted Monte Carlo sampling on filtered samples. The $\epsilon$ also controls the exploration similar to used in Reinforcement Learning\citep{sutton2018reinforcement}(refer algorithm \ref{alg:ew}). The value of $\epsilon$ reflects our belief in the accuracy of the failure predictor.  In the next section, we will empirically evaluate our approach with  all above mentioned acquisition functions: \textbf{top-B}, \textbf{DBAL-10 , DBAL-50}, \textbf{$\epsilon$-weighted Hybrid Query Strategy}, \textbf{Uniformly Random Monte Carlo}.     
\vspace*{-2mm}
\begin{algorithm}
\caption{$\epsilon$-weighted Sampling and de-biasing scheme}\label{alg:ew}
\begin{algorithmic}[1]
\State $Input: {trained\;teacher\;model(f),Design\; space(DS),Budget(B),\& \; \epsilon}$
\State $Output: {Candidate \; Samples(L_{candidate})}$
\State  $Initialization: S \leftarrow \{\}$
\For {$size(S) \leq \epsilon*B$} 
  \State $Sample\; proposal\; instance/ full\; DS, X_t$
  \State $Calculate\; f(X_t)$ \Comment{ Estimate failure probability on samples in DS} 
  \State $X_f \subseteq X_t : X_f = X_t\; \forall\; X_t, \;where \; f(X_t) \geq 0.5$ \Comment{ filtered high failure probability samples $X_f$}
  \State $X_{sel} = Choose\_Uniform\_rand(X_f , \epsilon*B)$ \Comment{ Choose $\epsilon*B$ samples randomly from $X_f$}
  \State $S \leftarrow S+X_{sel}$
\EndFor
\State $X_{rest}= Choose\_Uniform\_rand(X_t - X_{sel}, (1-\epsilon)*B)$ 
\State $S \leftarrow S+X_{rest}$
\end{algorithmic}
\end{algorithm}
\vspace*{-4mm}

\vspace*{-5mm}
\section{Related Work}
\vspace*{-3mm}
\label{sec:relatedWorks}
There are many ongoing and historical efforts for creating learning-based surrogate models in both propeller design and the FEA domain.\citep{vesting2014surrogate}\citep{vardhan2021machine}\citep{liang2018deep}.  However, without any strategic sampling, all work is done for limited design space.       
Active learning-based sampling by utilizing information-based objective function has also a long history\citep{fedorov2013theory}\citep{mackay1992information}\citep{settles2009active} and has been extensively used for regression problems under the Bayesian framework especially by using Kriging and Gaussian Process-based learning models\citep{forrester2008engineering}\citep{rasmussen2003gaussian}. With the advent of deep learning\citep{lecun2015deep}, it is found advantageous to integrate DL models(which is a powerful learning model) with AL to get the best of both worlds however the traditional AL literature has little to offer when the parameters of learning models are large. There is a plethora of ongoing research work in integrating DL with AL and it is called DeepAL but most work is done in the context of classification problems where direct probabilistic output makes creating information-based objective function much easier\citep{ren2021survey}. Our work is more inspired by \citep{uesato2018rigorous}\citep{vardhan2021rare}, in which an attempt made to learn the failure probability in the context of rare event failures in RL agents.

\vspace*{-5mm}
\section{Experiments and Evaluation}
\begin{figure}[!htb]
\minipage{0.43\textwidth}
  \includegraphics[width=1.2\textwidth]{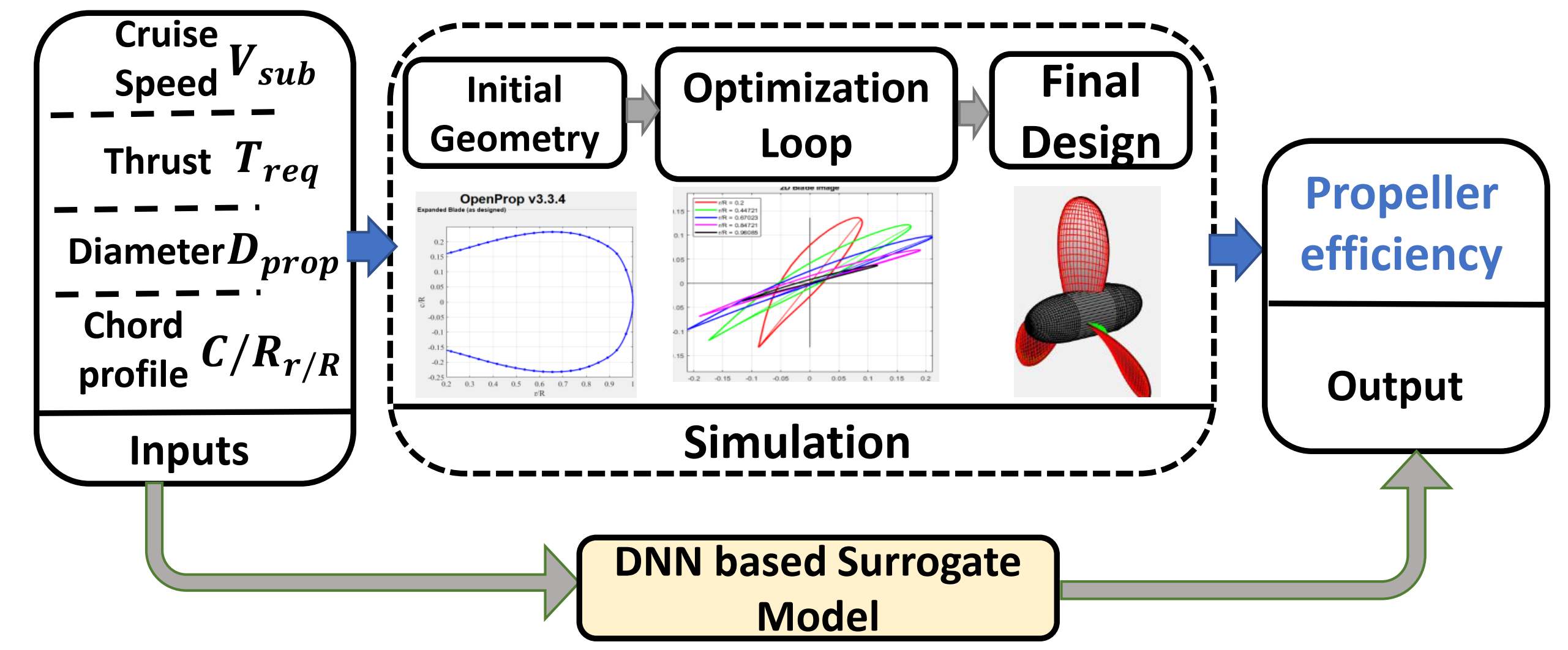}
   \vspace*{-7mm}
  \caption{Propeller design surrogate}\label{fig:prop_surr}
\endminipage\hspace{3.2em}%
\minipage{0.43\textwidth}
  \includegraphics[width=1.2\textwidth]{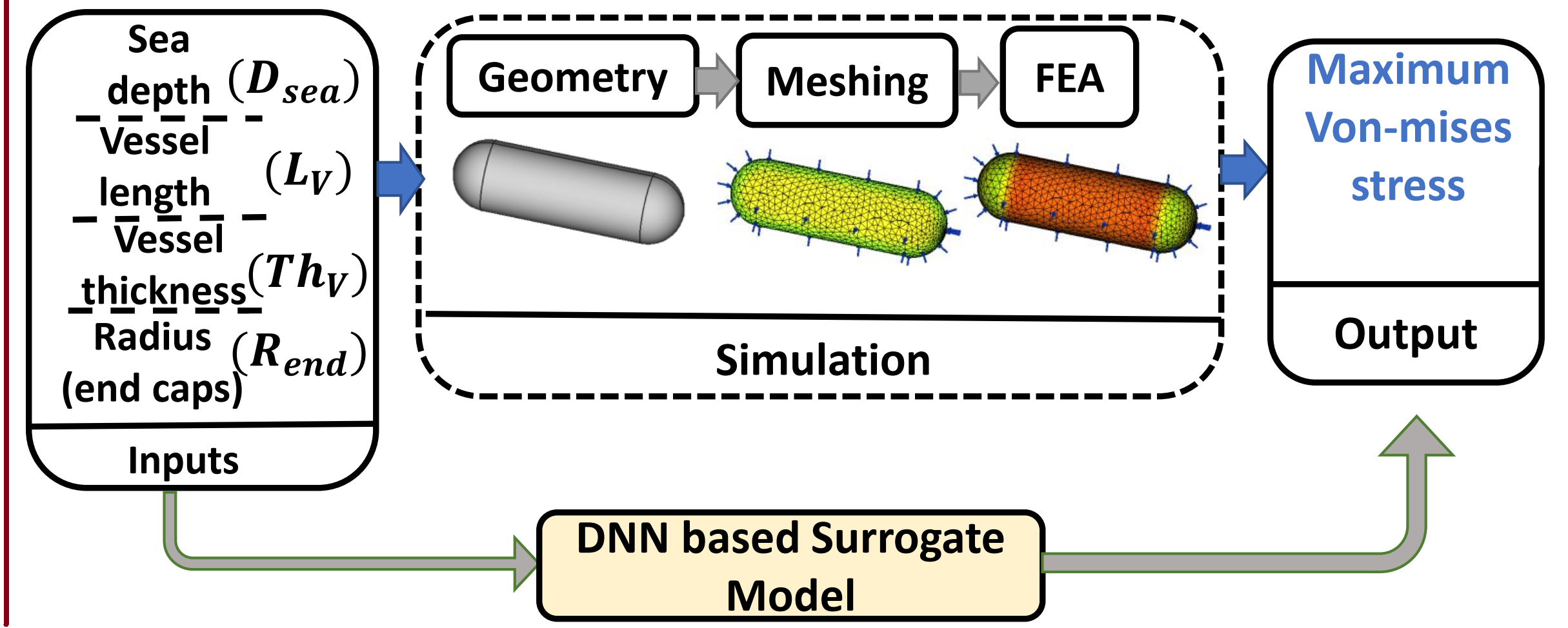}
  \vspace*{-7mm}
  \caption{Pressure vessel Surrogate}\label{fig:pv_surr}
\endminipage
\end{figure}

\vspace*{-3mm}
\label{sec:experiments}
For empirical evaluation, we select two engineering domains--propeller design domain and the Finite Element Analysis (FEA) domain. In both cases, we want to replace a physics-based simulation process with a computationally cheap DNN-based surrogate model. In the propeller domain, the goal is to create a surrogate model to predict the efficiency of the propeller on the given requirement and geometric design. In the case of the FEA domain, the goal is to create a surrogate model of the FEA process to predict the maximum Von mises stress (which determines the integrity of the vessel) in the capsule-shaped pressure vessel used in an Underwater vehicle (refer figures \ref{fig:prop_surr} and \ref{fig:pv_surr}).    

For data generation and evaluation, we used physics-based simulators -openprop \citep*{epps2009openprop} and freeCAD \citep*{riegel2016freecad} for the propeller and FEA domain respectively. In propeller cases, we generated 275000 labeled data in $14$ dimensional input parameters space because the average evaluation time per sample was small ($\approx\; 0.5\; sec$). However, in the case of pressure vessels, the evaluation time is high ($\approx\; 202\; sec$) and we generated 11131 labeled samples in the $4$ dimensional input parameter space. For experiment I in both domains, the total budget of training data is fixed to  2500 (50 iterations of the AL process, with a budget of 50 per iteration). The leftover labeled data is used as the test data (to measure the accuracy of the surrogate).  For evaluating the consistency of results, we ran multiple experiments and averaged the prediction accuracy.
  
\begin{figure}[!htb]
\minipage{0.37\textwidth}
  \includegraphics[width=1.2\textwidth]{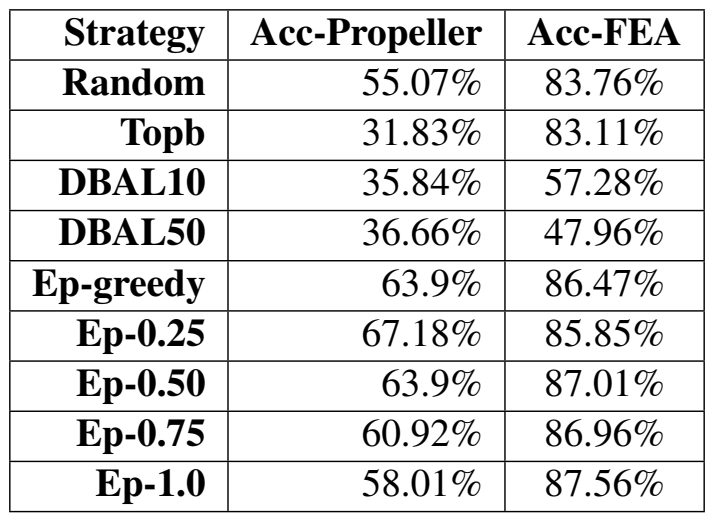}
   \vspace*{-10mm}
  \caption{Prediction accuracies}\label{fig:accuracy}
\endminipage\hspace{3.2em}%
\minipage{0.5\textwidth}
  \vspace*{-5mm}
  \includegraphics[width=1.2\textwidth]{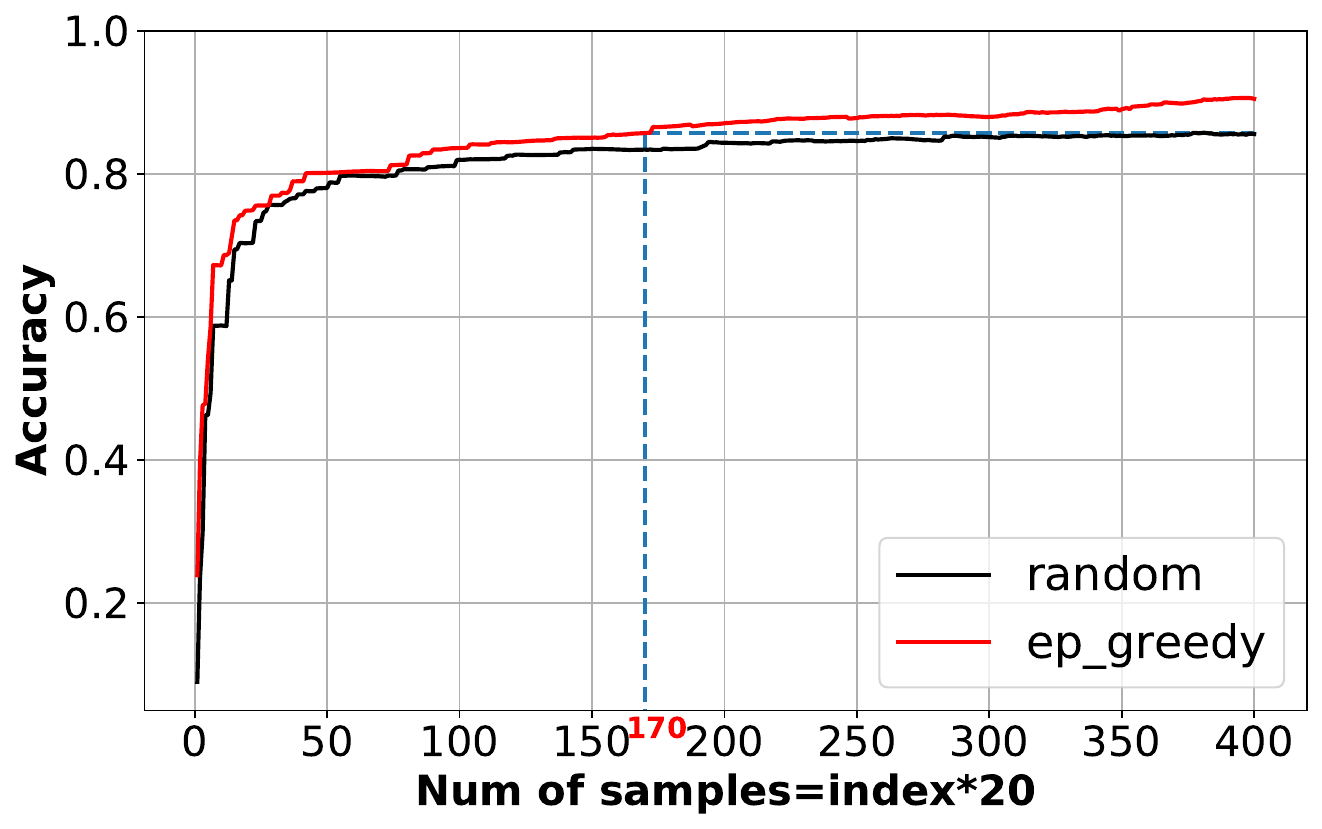}
  \vspace*{-10mm}
  \caption{FEA domain: Experiment II}\label{fig:free_res2}
\endminipage
\end{figure}

We observed that in both domains, \textbf{DBAL} and \textbf{top-b} strategies perform poorer than the random sampling and the underlying reason is the complete reliance on a biased estimation of failure probability.     
\begin{figure}[!htb]
\minipage{0.44\textwidth}
  \includegraphics[width=1.2\textwidth]{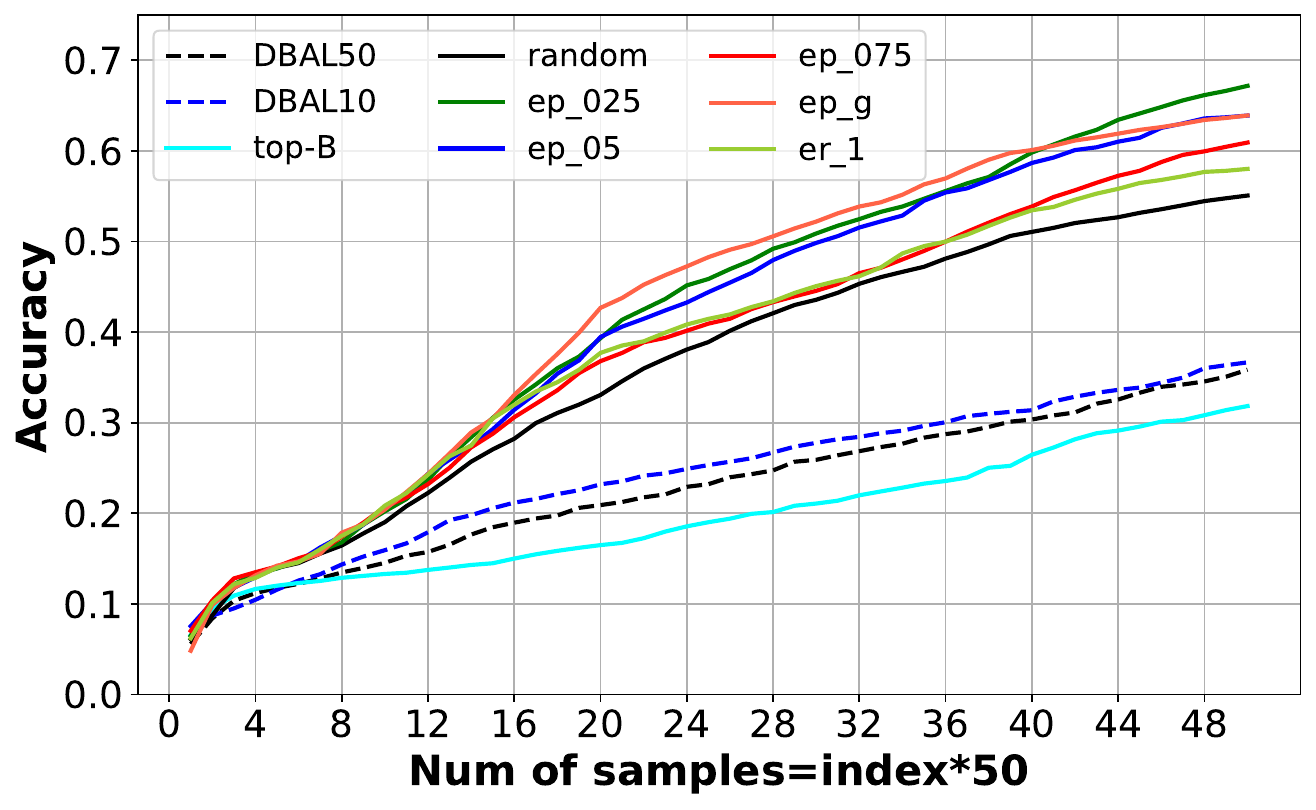}
   \vspace*{-10mm}
  \caption{Propeller Domain : Experiment I}\label{fig:prop_res1}
\endminipage\hspace{3.2em}%
\minipage{0.44\textwidth}
  \includegraphics[width=1.2\textwidth]{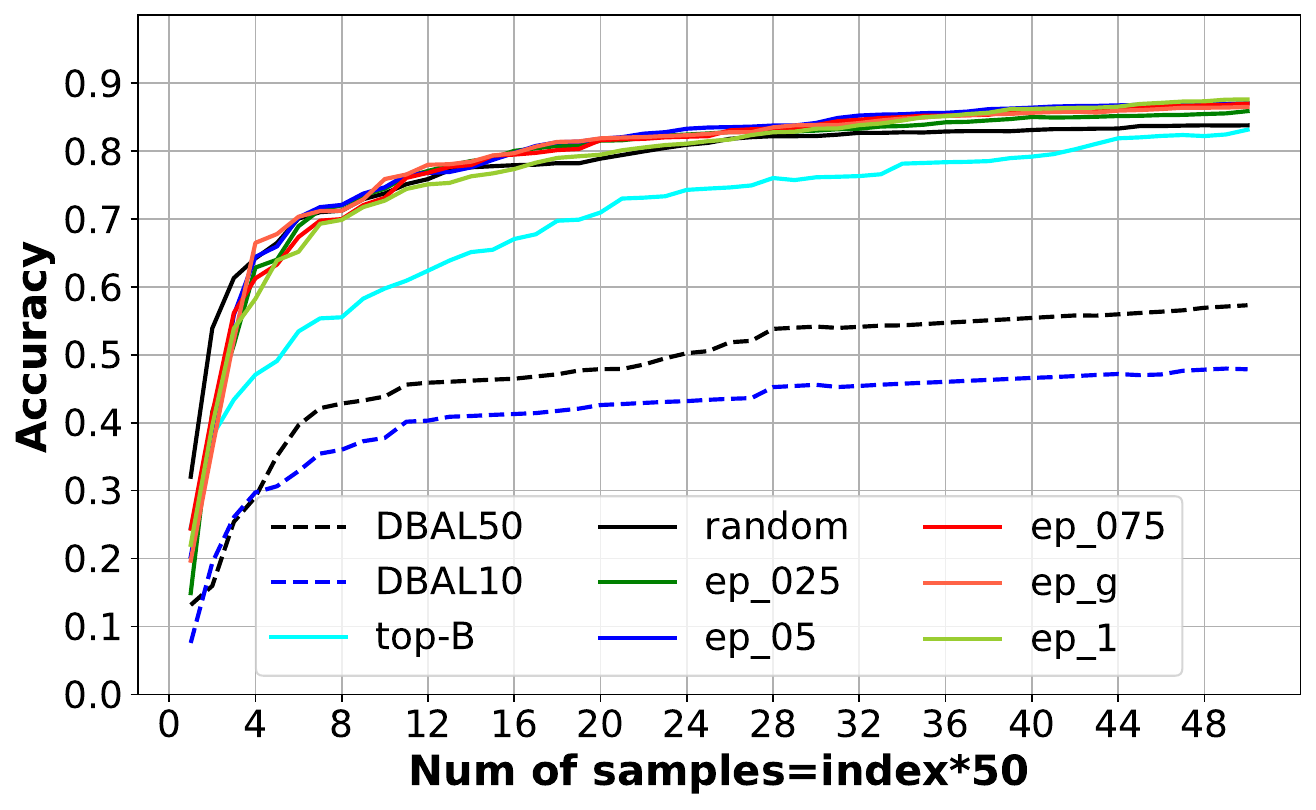}
  \vspace*{-10mm}
  \caption{FEA domain :Experiment II}\label{fig:free_res1}
\endminipage
\end{figure}
Figure \ref{fig:accuracy} shows the accuracy of trained DL-based surrogates using different sampling strategies (DL models and other learning parameters are kept the same). Our accuracy metrics is fraction of test predictions which are out of acceptable accuracy(+/- 5\% error) i.e., $\frac{\sum_{i=1}^{T} \mathbb{I}(|y_i-y_i^*| \leq 0.05*|y_i^*|)}{T} $. Here $y_i$ is a prediction on $i^{th}$ test sample having ground truth $y_i^*$ and $T$ is the total number of samples in the test data.  The \textbf{DBAL50} and \textbf{DBAL10} are two different DBAL (Diverse Batched Active learning) strategies with hyper-parameter $\beta=50$ and 10 respectively(refer \cite{zhdanov2019diverse}). In \textbf{random} strategy, at each iteration samples are selected uniformly randomly for unlabeled pool data($DS$). Five  $\epsilon-HQS$ approaches are tested labeled as $ep\_g,ep\_025,ep\_05,ep\_075$ and $ep\_1$, called as logarithmic increasing $\epsilon$, $\epsilon$ with weights 0.25,0.50, 0.75 and 1.0 respectively. The epsilon-greedy\citep{sutton2018reinforcement} approach increases belief in the prediction of teacher network logarithmic, while other fixed $\epsilon$ strategies assign linear belief to the teacher network.  For experiment II, we used only FEA domain and reduced the batch size per iteration to 20 and the increased the number of iterations to 400 (i.e., a total budget of training is increased from 2500 to 8000), and the rest of the data is used for testing. We observed the improvement in accuracy of surrogate using $epsilon$-greedy HQS at a higher rate than randomized sampling as the $epsilon$-greedy HQS can find strategic samples, which random sampling cannot. The accuracy increment was 4.14 \%(86.47\% to 90.61\%)  for $epsilon$-greedy HQS in comparison to 1.99\%(83.76\% to 85.75\%) in the case of random sampling when training data increased from 2500 samples to 8000 samples. Another result that we want to share is for reaching the same level of accuracy as the randomized sampling strategy, the $epsilon$-greedy HQS saves $57.5\%$ of samples and consequently $250$ hours of data generation time (includes the samples simulation time minus training and prediction time from teacher network).  

\vspace*{-4mm}
\vspace*{-5mm}
\section{Conclusion and Future Work}
\vspace*{-3mm}
\label{sec:conclusionFutureWork}
In this work, we developed a new method of data efficient surrogate modeling by combining Deep Learning and Active Learning for regression problem. When compared this methods like DBAL, VMC and top-B strategy, it outperformed all of them.

\vspace*{-5mm}
\section{Acknowledgments}
\vspace*{-3mm}
This work is supported by DARPA’s Symbiotic Design for CPS project (FA8750-20-C-0537).
\bibliography{references}
\end{document}